\documentclass[sigconf,nonacm]{acmart}

\usepackage{mdframed}    
\usepackage{appendix}    
\usepackage{threeparttable}
\usepackage{algorithm}
\usepackage{algorithmic}
\usepackage{booktabs}
\usepackage{enumitem}
\usepackage{multirow}
\usepackage{multicol}
\usepackage{array}
\usepackage{pifont}
\usepackage{listings}
\usepackage{amsmath}
\usepackage{color}
\usepackage{xcolor}
\usepackage{colortbl}

\AtBeginDocument{%
  }

\setcopyright{acmlicensed}
\copyrightyear{2018}
\acmYear{2018}
\acmDOI{XXXXXXX.XXXXXXX}

\acmConference[Conference acronym 'XX]{Make sure to enter the correct
  conference title from your rights confirmation emai}{June 03--05,
  2018}{Woodstock, NY}
\acmISBN{978-1-4503-XXXX-X/18/06}

\begin{document}

\title{ChineseEcomQA: A Scalable E-commerce Concept Evaluation Benchmark for Large Language Models}



\author{Haibin Chen$^{\dagger}$}
\thanks{$^{\dagger}$Both authors contributed equally to this research.}
\affiliation{%
  \institution{Taobao \& Tmall Group of Alibaba}
  \city{Hangzhou}
  \country{China}}

\author{Kangtao Lv$^{\dagger}$}
\affiliation{%
  \institution{Taobao \& Tmall Group of Alibaba}
  \city{Hangzhou}
  \country{China}}

\author{Chengwei Hu}
\affiliation{%
  \institution{Taobao \& Tmall Group of Alibaba}
  \city{Hangzhou}
  \country{China}}

\author{Yanshi Li}
\affiliation{%
  \institution{Taobao \& Tmall Group of Alibaba}
  \city{Beijing}
  \country{China}}

\author{Yujin Yuan}
\affiliation{%
  \institution{Taobao \& Tmall Group of Alibaba}
  \city{Hangzhou}
  \country{China}}
  
\author{Yancheng He}
\affiliation{%
  \institution{Taobao \& Tmall Group of Alibaba}
  \city{Beijing}
  \country{China}}

\author{Xingyao Zhang}
\affiliation{%
  \institution{Taobao \& Tmall Group of Alibaba}
  \city{Hangzhou}
  \country{China}}
  
\author{Langming Liu}
\affiliation{%
  \institution{Taobao \& Tmall Group of Alibaba}
  \city{Hangzhou}
  \country{China}}
  
\author{Shilei Liu}
\affiliation{%
  \institution{Taobao \& Tmall Group of Alibaba}
  \city{Hangzhou}
  \country{China}}
  
\author{Wenbo Su}
\affiliation{%
  \institution{Taobao \& Tmall Group of Alibaba}
  \city{Beijing}
  \country{China}}

\author{Bo Zheng}
\affiliation{%
  \institution{Taobao \& Tmall Group of Alibaba}
  \city{Beijing}
  \country{China}}

\renewcommand{\shortauthors}{Trovato et al.}

\begin{abstract}
With the increasing use of Large Language Models (LLMs) in fields such as e-commerce, domain-specific concept evaluation benchmarks are crucial for assessing their domain capabilities. Existing LLMs may generate factually incorrect information within the complex e-commerce applications. Therefore, it is necessary to build an e-commerce concept benchmark. Existing benchmarks encounter two primary challenges: (1) handle the heterogeneous and diverse nature of tasks, (2) distinguish between generality and specificity within the e-commerce field.
To address these problems, we propose \textbf{ChineseEcomQA}, a scalable question-answering benchmark focused on fundamental e-commerce concepts. ChineseEcomQA is built on three core characteristics: \textbf{Focus on Fundamental Concept}, \textbf{E-commerce Generality} and \textbf{E-commerce Expertise}. Fundamental concepts are designed to be applicable across a diverse array of e-commerce tasks, thus addressing the challenge of heterogeneity and diversity. Additionally, by carefully balancing generality and specificity, ChineseEcomQA effectively differentiates between broad e-commerce concepts, allowing for precise validation of domain capabilities.
We achieve this through a scalable benchmark construction process that combines LLM validation, Retrieval-Augmented Generation (RAG) validation, and rigorous manual annotation. Based on ChineseEcomQA, we conduct extensive evaluations on mainstream LLMs and provide some valuable insights. We hope that ChineseEcomQA could guide future domain-specific evaluations, and facilitate broader LLM adoption in e-commerce applications.

\end{abstract}



\maketitle

\section{Introduction}
Recent years have witnessed rapid advancements in large language models (LLMs) and their widespread adoption in different fields such as e-commerce. In the e-commerce domain, accurate comprehension of fundamental concepts is a crucial prerequisite for successful application of LLMs~\cite{Yu2024}. However, existing LLMs often suffer from generating lengthy responses that contain factual inaccuracies (hallucination)~\cite{huang2025survey}, making it challenging to systematically assess their factual abilities~\cite{he2024chinese}. Therefore, it is necessary to build a scalable e-commerce knowledge benchmark.

\begin{figure*}[t]
    \centering
    \includegraphics[width=0.9\linewidth]{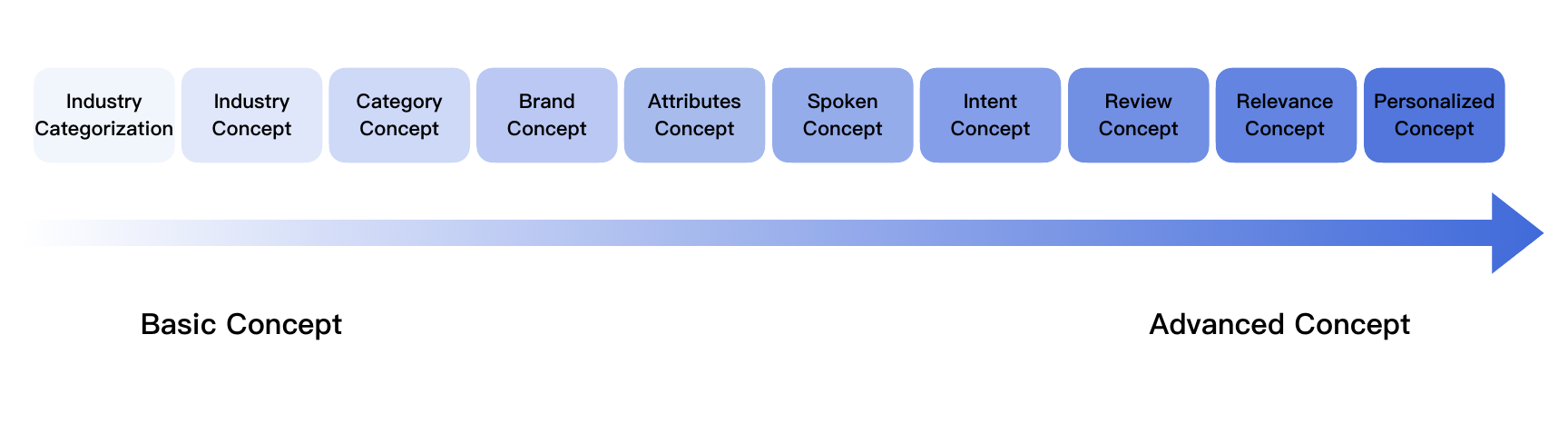}
    \caption{Overview of fundamental e-commerce concepts. From basic concept to advanced concept, we categorize into 10 sub-concepts.
    }
    \label{fig:overview}
\end{figure*}

Constructing scalable evaluation benchmarks faces two specific challenges: (1) Heterogeneity and Diverse
Nature: The e-commerce domain encompasses a wide range of task formats, with definitions varying significantly across scenarios ~\cite{jin2024shopping}. For example, understanding user queries includes query error correction, query tagging, and other processes. (2)  Distinguish between Generality and Specificity: While e-commerce knowledge intersects with general world knowledge, it requires a high degree of specialized expertise. Addressing practical e-commerce problems necessitates the integration of domain-specific knowledge with general knowledge. Based on key features of existing factuality benchmarks (such as high quality and static) , we argue that a scalable e-commerce benchmark should possess three essential characteristics:

\begin{itemize}[leftmargin=*]
\item \textbf{Focus on Fundamental Concept}: We focus on fundamental concepts that enable unified generative evaluation.

\item \textbf{E-commerce Generality}: The concepts assessed by the benchmark must be common across the e-commerce industry, avoiding platform-specific implementations or task-specific formulations.

\item \textbf{E-commerce Expertise}: Real-world e-commerce problems often require a foundation of specialized e-commerce knowledge, complemented by the application of general comprehension and reasoning skills.

\end{itemize}

Existing works like ShoppingMMLU~\cite{jin2024shopping} proposed diversified e-commerce instruction evaluation. Their common idea is to abstract multiple e-commerce tasks into several main task skills. However, they lack the granularity to dissect and analyze specific, foundational e-commerce concepts. Besides, they lack a comprehensive process to ensure both the e-commerce generality and e-commerce expertise. To address this gap, we propose a highly scalable dataset construction process that combines the strengths of LLM validation, Retrieval-Augmented Generation (RAG) validation, and rigorous manual annotation. This hybrid process ensures the coverage of three core characteristics outlined above. Fundamental concepts are designed to be applicable across a diverse array of e-commerce tasks, thus addressing the challenge of heterogeneity and diversity. Additionally, by carefully balancing generality and specificity, ChineseEcomQA~\footnote{~\url{https://github.com/OpenStellarTeam/ChineseEcomQA}} effectively differentiates between broad e-commerce concepts, allowing for precise validation of domain capabilities. The resulting benchmark, ChineseEcomQA, encompasses 20 major industries and 10 core concept dimensions, comprising 1,800 carefully curated question-answer pair. Through extensive evaluations of mainstream LLMs using ChineseEcomQA, we reveal several key insights:
\begin{itemize}[leftmargin=*]

\item \textit{Leading Models}: Deepseek-R1 and Deepseek-V3 are currently the best models, demonstrating the promising potential of powerful foundation LLMs (and reasoning LLMs) in the e-commerce field.

\item \textit{Significant Challenges}: ChineseEcomQA poses considerable challenges, with many state-of-the-art models achieving below 60\% accuracy on specific sub-concepts.

\item \textit{Scaling Laws}: E-commerce concepts follow scaling law, where larger models demonstrate superior capability in advanced concepts.

\item \textit{Calibration}: Larger models show better calibration in confidence estimation.

\item \textit{Reasoning LLMs}: Deepseek-R1-Distill-Qwen series performs worse than the original Qwen series and struggles to identify and correct its own factual errors, indicating that there are still many challenges in the reasoning ability of open domains.

\item \textit{RAG matters}: When introducing the RAG strategy into existing LLMs, models of various sizes have shown significant performance improvements, narrowing the gap among models.

\end{itemize}

The main contributions to our work are threefold:

\begin{itemize}[leftmargin=*]

\item We define and categorize core e-commerce concepts that are scalable to different e-commerce tasks. A standardized QA evaluation benchmark has been constructed around e-commerce core concepts.

\item We propose a highly scalable domain-specific benchmark construction process, ensuring that ChineseEcomQA adheres three core characteristics (Focus on Fundamental Concept, E-commerce Generality, E-commerce Expertise).

\item We conduct thorough experiments on existing LLMs, providing valuable insights into their strengths and weaknesses, and offering directions for future research in the e-commerce field.

\end{itemize}


\begin{figure*}[t]
    \centering
    \includegraphics[width=0.95\linewidth]{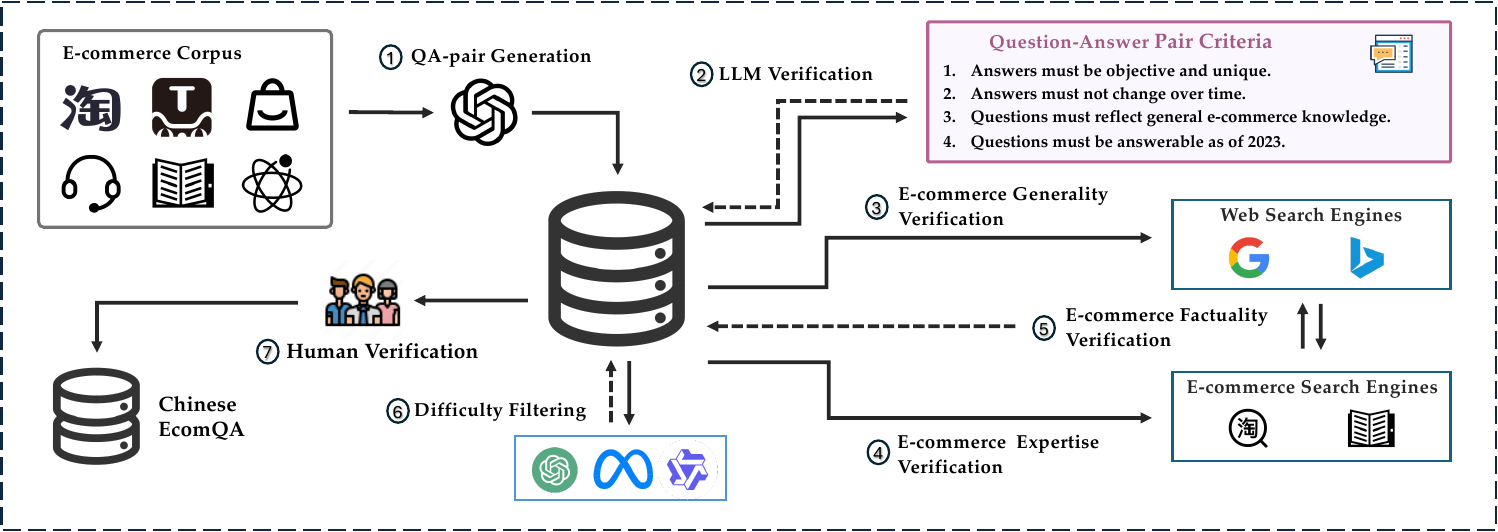}
    \caption{
    An overview of the data construction process of ChineseEcomQA.
    }
    \label{fig:data_pipeline}
\end{figure*}

\begin{figure*}[t]
    \centering
    \includegraphics[width=0.95\linewidth]{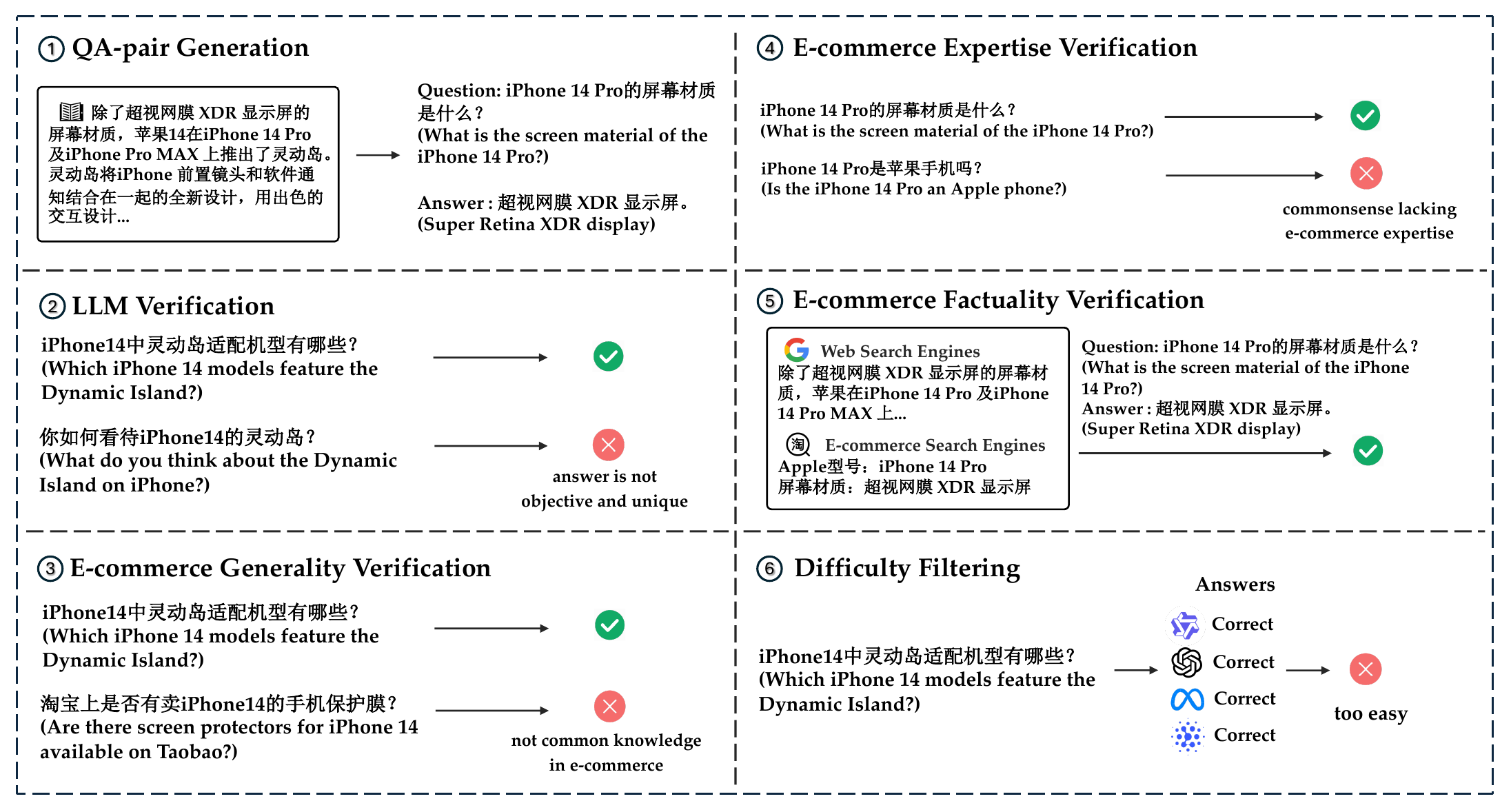}
    \caption{
    Illustrative examples of the data construction process.
    }
\label{fig:data_pipeline_example}
\end{figure*}

\section{ChineseEcomQA}
\subsection{Overview}
Figure \ref{fig:overview} shows the fundamental e-commerce
concepts. Figure \ref{fig:data_pipeline} shows the overview of the data construction process of ChineseEcomQA. Besides, we provide illustrative examples of the data construction process in Figure \ref{fig:data_pipeline_example}. Table \ref{tab:dataset_statistics} shows the statistics of ChineseEcomQA. Figure \ref{fig:model_radar} visualizes the results of some selected models on ten sub concept tasks. In Appendix ~\ref{sec:appendix A}, we provide some examples of ChineseEcomQA. In the following subsection, we will introduce fundamental e-commerce concepts, data construction process, dataset statistics and evaluation metrics.

\subsection{Fundamental E-commerce Concepts}
Starting from the basic elements of e-commerce such as user behavior and product information, we summarized the main types of e-commerce concepts, defined 10 sub-concepts from basic concepts to advanced concepts as follows:

\begin{itemize}[leftmargin=*]

\item \textbf{Industry Categorization.} Given e-commerce corpus (such as user queries or web corpus), the LLMs need to figure out which e-commerce industries and categories are involved. The difficulty lies in distinguishing similar categories in the e-commerce domain.

\item \textbf{Industry Concept.} The model needs to understand the specialized knowledge in different e-commerce industries. The difficulty lies in accurately memorizing professional factual knowledge.

\item \textbf{Category Concept.} The model must understand which category a common, standard product belongs to.

\item \textbf{Brand Concept.} The model needs to recognize major brands and understand some background information about them.

\item \textbf{Attribute Concept.} E-commerce text often describes products using basic attributes, like style or age group. The model must have the ability to pick out these specific attribute words.

\item \textbf{Spoken Concept.} The e-commerce field is closely related to daily life scenarios, and people often use casual and imprecise language to express what they want. The model needs to understand the true expression forms.

\item \textbf{Intent Concept.} Beyond just informal language, sometimes consumers just list a bunch of attributes. The model needs to figure out the consumer's true intention from these phrases (such as how to choose).

\item \textbf{Review Concept.} The model needs to understand common concepts in user comments, such as emotional tendencies, commonly used evaluation aspects, etc.

\item \textbf{Relevance Concept.} One of the most crucial concepts of e-commerce is figuring out how relevant a product is to what a user wants. The model needs to integrate basic concepts such as intent concept and category concept to determine the relevance among user expression and products.

\item \textbf{Personalized Concept.} Personalized concept is one of the most important parts of user experience. This requires combining basic e-commerce concepts with general reasoning skills to recommend new product categories that best match a user's recent preferences.

\end{itemize}

\subsection{Data Collection}
\subsubsection{QA-pair Generation}
We collect a large amount of knowledge-rich e-commerce corpus, rich in information and covering various related concepts. Then, we prompt the LLM (GPT-4o) to generate question-answer pairs faithfully based on the given contents. For more open questions, we require the LLM to simultaneously provide candidate answers that are highly confusing and difficult. Providing candidate options in some concepts is beneficial for the objectivity and uniqueness of the evaluation.

\subsubsection{LLM Verification}
In the previous subsection, we collected a large number of question-answer pairs. To ensure the basic quality of dataset, we use LLM (GPT-4o) to filter data that cannot meet the requirements of our predefined criteria. Specifically, the question-answer pairs must meet the following criteria.

\begin{itemize}[leftmargin=*]
\item Questions must be a clear question about an e-commerce concept. There should be no ambiguity in the problem statement. For example, "What is the most well-known brand of washing machine?" is a disqualification question, because "most familiar" may be controversial.

\item Questions must be answerable as of 2023. The e-commerce concept investigated in the question cannot use the industry knowledge after December 31, 2023.

\item Answers must be objective and unique. There should be only one clear and objective answer to the question raised. For example, "What do you think about the Dynamic Islang on iPhone? " is too subjective.

\item Answers must not change over time. Questions about current trends or the latest product series, which are constantly evolving, are not suitable.

\end{itemize}

\subsubsection{E-commerce Generality Verification}
Since we use e-commerce corpus to construct the question-answering pairs, the original data may contain platform-specific knowledge, which affects the generality of the dataset. Therefore, we deploy external retrieval tools (i.e., web search engines) to gather information from a wider range of sources. This helped the LLM to evaluate the generality of dataset. For some question types, like those requiring the model to choose the correct category from a list, direct web searching wasn't practical. In these cases, we first require the LLM to identify several knowledge points related to the question. We then used these knowledge points as search queries. After getting the search results, we used the LLM again to remove any data that does not conform to domain generality. As illustrated in Figure \ref{fig:data_pipeline_example}, "Are there screen protectors for iPhone 14 available on Taobao?" is not common knowledge in e-commerce.

\subsubsection{E-commerce Expertise Verification}
While web searches helped ensure e-commerce generality, we also needed to confirm that the questions had sufficient depth within the e-commerce domain. Therefore, we use e-commerce search engines (such as Taobao search and e-commerce encyclopedia) to obtain more specialized information. We require the LLM to judge the level of expertise, as concept that is too basic will be filtered out.

\subsubsection{E-commerce Expertise Verification}
Real-world e-commerce problems require to integrate domain expertise and general knowledge. To verify the factual correctness of question-answer pairs, we require the LLM consider information from both general web searches and specialized e-commerce searches. For example in Figure \ref{fig:data_pipeline_example}, regarding the question of iPhone 14 Pro, we used web search results and product encyclopedias to comprehensively determine the correctness of the facts.

\subsubsection{Difficulty Filtering}
We discover the knowledge boundaries of the LLMs by checking whether multiple LLMs answer correctly. Specifically, we selected models from the Qwen series, LLaMA series, and GPT-4o for evaluation. If all the models answer a question correctly, we considered it too easy and removed it.

\subsubsection{Human Verification}
Finally, we use manual annotation to verify the quality of the evaluation set. Manual annotation requires a comprehensive consideration of the characteristics mentioned above, such as domain generality, domain expertise, and overall quality. In fact, most of the problematic data has already been filtered by the previous process. Combining LLM's ability to integrate information, verify information, and perform manual verification, we believe it is a more scalable construction process.

\subsection{Dataset Statistics}
\begin{figure}[t]
    \centering
    \includegraphics[width=0.95\linewidth]{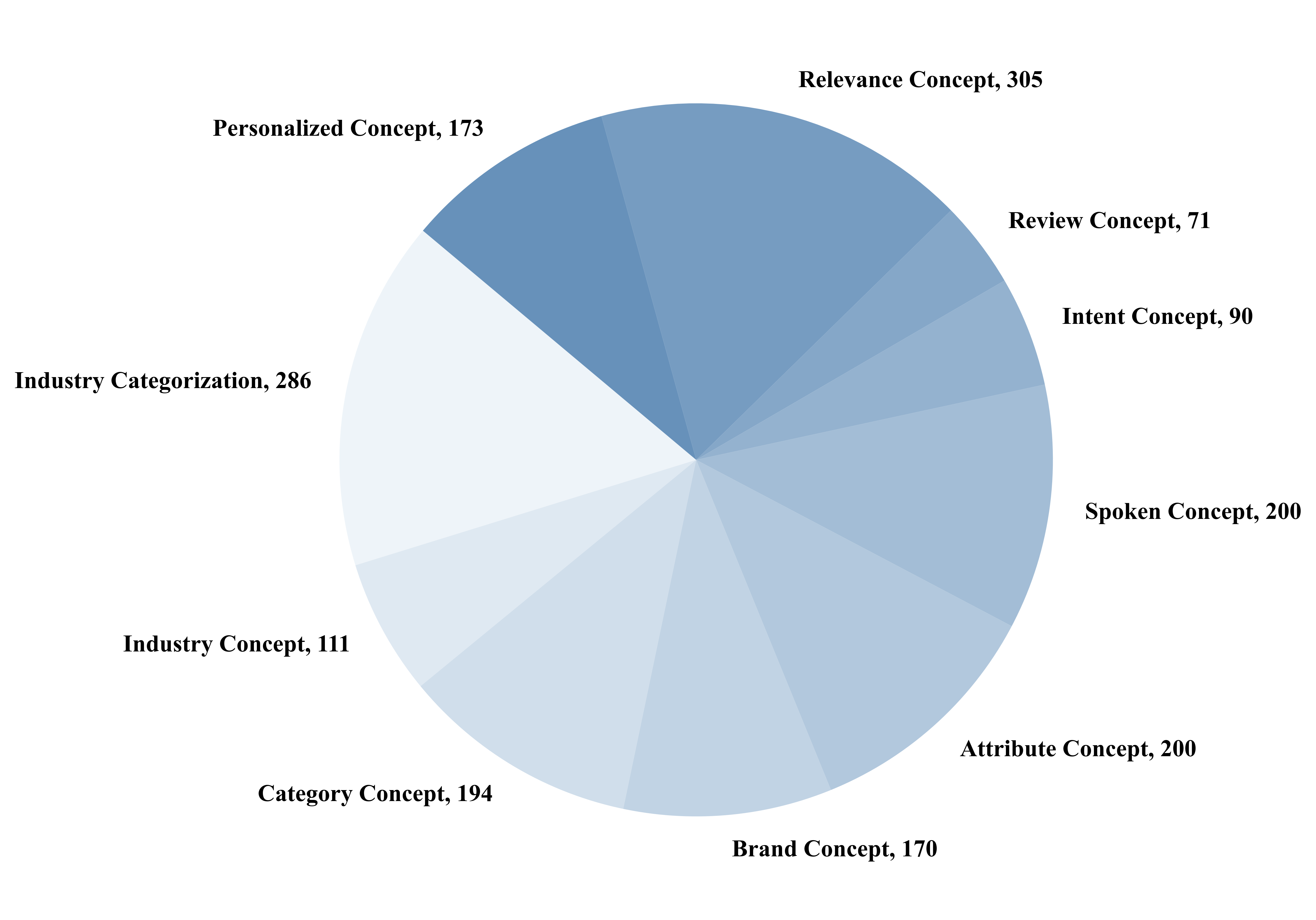}
    \caption{
    Dataset statistics of ChineseEcomQA.
    }
    \label{tab:dataset_statistics}
\end{figure}

Figure \ref{tab:dataset_statistics} presents the statistics of ChineseEcomQA. With a total of 1,800 samples, ChineseSimpleQA \cite{he2024chinesesimpleqachinesefactuality} relatively evenly distributed 10 sub-concept types. Furthermore, the average length of reference answers is 18.26. The concise and unified format, consistent with the articles in the SimpleQA series~\cite{wei2024measuring,he2024chinese}, has the advantages of easy-to-evaluate and relatively low evaluation cost.


\subsection{Evaluation Metrics}
Given reference answer, LLM-as-a-judge is an effective and prevalence method \cite{gu2024survey}. Based on the questions, candidate answers, and reference answers, we use GPT-4o, Claude-3.5-Sonnet and Deepseek-V3 \cite{liu2024deepseek} as the judge models. The final judgment is determined by the voting results of three LLMs. There are three types of evaluation criteria: (1) Correct: The candidate answer fully includes the reference answer without introducing any contradictory elements. (2) Wrong: The candidate answer contains factual statements that contradict the reference answer. (3) Not Attempted: The LLM is not confident enough to provide specific answers, and there are no statements in the response that contradict the reference answer. The evaluation prompt can be found in the Appendix \ref{sec:appendix B}. In the results section that follows, we use the proportion of "Correct" answers as our accuracy metric.

\section{Experiments}
\begin{table*}[ht]
    \centering
    \begin{tabular}{c|c|cccccccccc}
        \toprule
         & \textbf{Accuracy} & \multicolumn{10}{c}{\textbf{Accuracy on 10 sub-concepts}} \\
        \cmidrule(lr){2-12}
         \textbf{Models} & Avg. &IC & IDC & CC & BC & AC & SC & ITC & RVC & RLC & PC \\
        \cmidrule(lr){1-12}
        \rowcolor{lightgray} \multicolumn{12}{c}{{\bf \textit{Closed-Source Large Language Models}}} \\
        \cmidrule(lr){1-12}
        GLM-4-Plus & 69.2 & 57.3 & 54.1 & 76.3 & 77.6 & 69.5 & 59.5 & 72.2 & 83.1 & 68.5 & 74.0  \\
        Qwen2.5-max & 68.5 & 62.2 & 62.2 & 71.1 & 77.6 & 63.0 & 58.5 & 57.8 & 88.7 & 63.9 & 80.4 \\
        Yi-Large & 67.6 & 56.6 & 59.5 & 71.1 & 81.8 & 62.0 & 58.5 & 70.0 & 70.4 & 68.5 & 78.0 \\
        o1-preview & 66.8 & 69.2 & 63.1 & 78.4 & 80.0 & 67.0 & 52.0 & 43.3 & 83.1 & 61.3 & 71.1 \\
        Baichuan4-Turbo & 66.4 & 57.3 & 56.8 & 82.0 & 72.4 & 61.0 & 59.5 & 66.7 & 78.9 & 55.4 & 74.6 \\
        GPT-4o & 65.6 & 68.2 & 52.3 & 74.7 & 72.4 & 64.5 & 56.5 & 50.0 & 80.3 & 57.7 & 79.8 \\
        Doubao-1.5-pro-32k & 64.0 & 69.6 & 64.0 & 62.9 & 74.1 & 56.5 & 64.5 & 48.9 & 69.0 & 62.6 & 68.2 \\
        Claude-3.5-Sonnet & 63.8 & 70.6 & 56.8 & 73.2 & 64.1 & 63.0 & 31.5 & 62.2 & 81.7 & 65.2 & 69.4  \\
        Gemini-1.5-pro & 61.1 & 59.8 & 49.6 & 67.0 & 70.0 & 56.0 & 43.5 & 55.6 & 81.7 & 54.1 & 73.4 \\
        o1-mini & 55.4 & 59.1 & 41.4 & 53.1 & 37.1 & 59.0 & 53.0 & 58.9 & 64.8 & 63.6 & 64.2  \\
        Gemini-1.5-flash & 54.5 & 62.9 & 35.1 & 57.2 & 46.5 & 52.5 & 53.0 & 36.7 & 74.7 & 54.4 & 71.7 \\
        \cmidrule(lr){1-12}
        \rowcolor{lightgray} \multicolumn{12}{c}{{\bf \textit{Open-Source Large Language Models}}} \\
        \cmidrule(lr){1-12}
        DeepSeek-R1 & 74.0 & 62.9  & 72.1 & 72.1 & 84.7  & 70.5 & 55.5 & 67.8 & 85.9 & 76.1 & 92.5 \\
        DeepSeek-V3 & 72.2 & 67.5 & 64.9 & 74.2 & 80.6 & 69.0 & 62.0 & 72.2 & 77.5 & 68.2 & 86.1 \\
        DeepSeek-V2.5 & 67.4 & 66.4  & 58.6  & 73.7 & 76.5  & 64.0 & 60.0 & 75.6 & 83.1 & 54.1 & 61.8 \\
        DeepSeek-67B & 58.4 & 61.2  & 47.7  & 70.6 & 62.9  & 47.0 & 52.5 & 60.0 & 59.2 & 55.7 & 67.1 \\
        DeepSeek-7B & 47.5 & 38.5  & 41.1  & 59.3 & 45.9  & 40.0 & 49.0 & 54.4 & 47.9 & 54.4 & 44.7 \\
        \cmidrule(lr){1-12}
        DeepSeek-R1-Distill-Qwen-32B & 57.1 & 63.6 & 46.0 & 62.4 & 47.6 & 36.0 & 43.0 & 61.1 & 78.9 & 61.6 & 70.6 \\
        DeepSeek-R1-Distill-Qwen-14B & 50.6 & 64.7 & 43.2 & 62.9 & 38.8 & 27.5 & 41.0 & 60.0 & 67.6 & 59.0 & 40.9 \\
        DeepSeek-R1-Distill-Qwen-7B & 38.9 & 48.6 & 18.0 & 53.6 & 16.5 & 25.0 & 33.5 & 36.7 & 52.1 & 48.2 & 57.2 \\
        DeepSeek-R1-Distill-Qwen-1.5B & 26.2 & 35.7 & 2.7 & 46.9 & 6.5 & 8.5 & 23.5 & 18.9 & 40.9 & 40.0 & 38.2 \\
        \cmidrule(lr){1-12}
        Qwen2.5-72B & 62.7 & 57.3  & 46.0  & 66.0  & 64.7  & 55.5  & 58.0  & 67.8  & 76.1  & 56.7  & 78.6 \\
        Qwen2.5-32B & 60.9 & 62.2  & 42.3  & 58.8  & 50.6  & 61.5  & 52.5  & 66.7  & 74.7  & 62.3  & 77.5 \\
        Qwen2.5-14B & 55.3 & 57.0  & 40.5  & 54.6  & 48.8  & 59.0  & 49.0  & 40.0  & 66.2  & 59.3  & 78.6  \\
        Qwen2.5-7B & 47.1 & 45.8  & 24.3  & 51.6  & 37.6  & 44.5  & 54.0  & 31.1  & 64.8  & 48.5  & 68.8 \\
        Qwen2.5-3B & 41.7 & 52.1  & 14.4  & 41.8  & 34.1  & 42.5  & 34.0  & 30.0  & 60.6  & 51.1  & 56.7 \\
        \cmidrule(lr){1-12}
        LLaMA3.1-70B & 54.6 & 59.1 & 35.7 & 58.8 & 39.4 & 58.0 & 37.5 & 73.3 & 74.7 & 53.8 & 56.1 \\
        LLaMA3.1-8B & 42.4 & 40.6 & 11.7 & 61.3 & 17.1 & 42.0 & 38.5 & 42.2 & 66.2 & 44.6 & 60.0 \\
        \bottomrule
        \bottomrule
    \end{tabular}
    \vspace{5mm}
    \caption{Results of different models on ChineseEcomQA. For
sub-concepts, IC, IDC, CC, BC, AC, SC, ITC, RVC, RLC and PC represent “Industry Categorization”, “Industry Concept”, “Category Concept”, “Brand Concept”, “Attribute Concept”, “Spoken Concept”, “Intent Concept”, “Review Concept”, “Relevance Concept” and “Personalized Concept” respectively.}
    \label{tab:main_result}
    \vspace{-3mm}
\end{table*}

\subsection{Baseline Models}
We evaluate 11 closed-source LLMs (i.e., o1-preview, Qwen2.5-max~\cite{yang2024qwen2}, Doubao-1.5-pro-32k, GLM-4-Plus, GPT-4o~\cite{OpenAI2023GPT4}, Gemini-1.5-pro~\cite{team2024gemini}, Gemini-1.5-flash~\cite{team2024gemini}, Claude-3.5-Sonnet, Yi-Large, Baichuan4-Turbo, o1-mini), and 16 open-source LLMs (i.e., Qwen2.5 series~\cite{yang2024qwen2}, LLaMA3.1 series~\cite{dubey2024llama}, DeepSeek series~\cite{bi2024deepseek,liu2024deepseek,guo2025deepseek}).

\subsection{Main Results}

As shown in Table \ref{tab:main_result}, we provide the performance results of different LLMs on our ChineseEcomQA. From Table \ref{tab:main_result} and Figure \ref{fig:model_radar}, we have the following insightful observations: 

\begin{figure*}[t]
    \centering
    \includegraphics[width=1\linewidth]{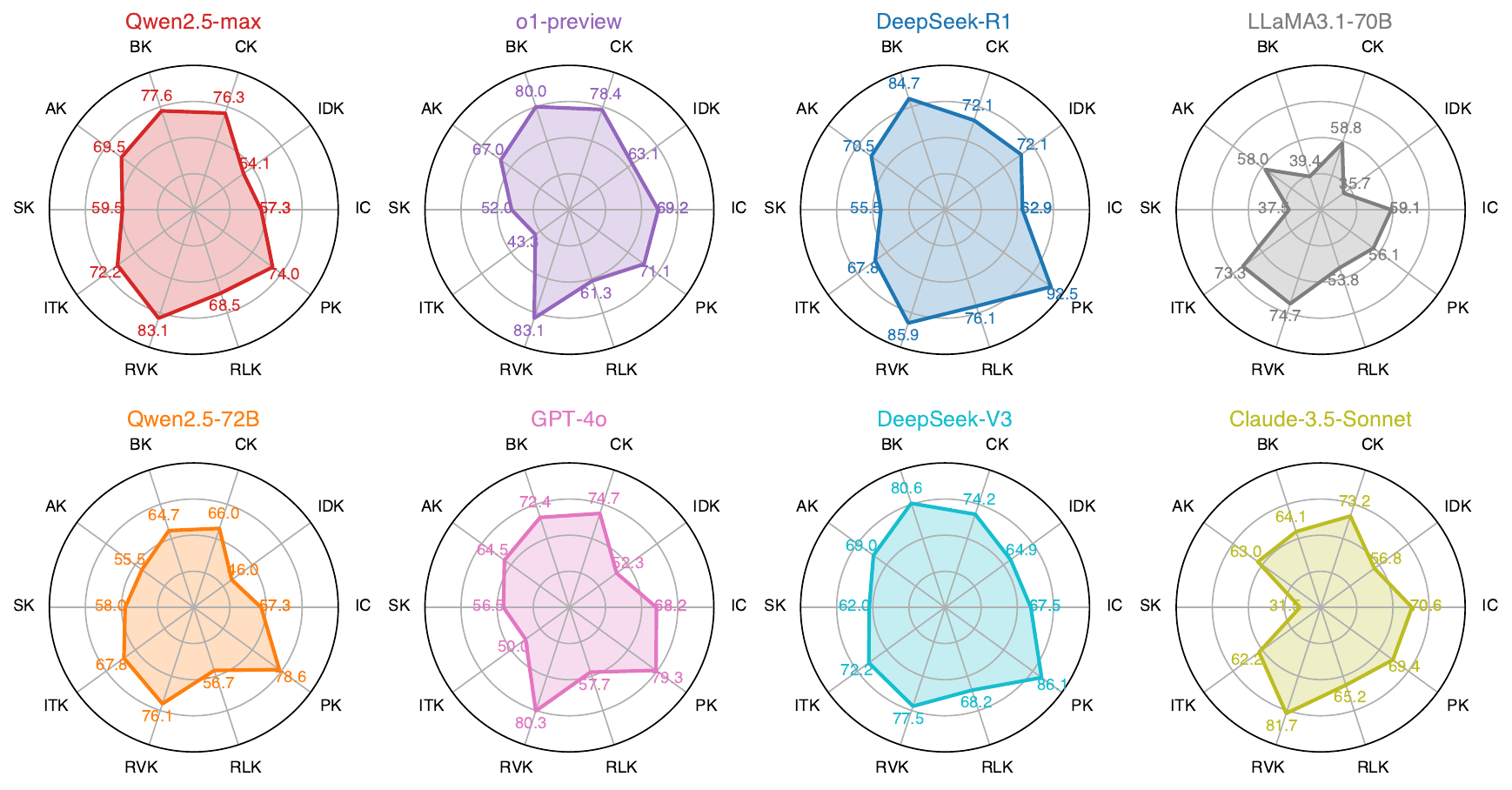}
    \caption{
    Detailed results on some selected models across ten sub-concept tasks.
    }
    \label{fig:model_radar}
\end{figure*}

\begin{itemize}[leftmargin=*]
\item Deepseek-R1 \cite{guo2025deepseek} and Deepseek-V3 \cite{liu2024deepseek} are currently the best models, demonstrating the promising potential of powerful foundation LLMs (reasoning LLMs) in the e-commerce field.

\item The conclusion of Scaling Law holds true. There is significant overlap between concepts in e-commerce and general knowledge. For some complex e-commerce concepts, the performance difference between larger and smaller LLMs is more pronounced.

\item LLMs developed by Chinese companies generally show better performance, especially on advanced e-commerce concept. While O1-preview perform better on basic concepts, it faces difficulties with more advanced ones.

\item Certain types of e-commerce concepts, such as relevance concept, still pose significant challenges for LLMs. Large models, with their strong general capabilities, can partially generalize to e-commerce tasks, whereas smaller models struggle significantly in this domain. These characteristics underscore the necessity of developing models specifically tailored for the e-commerce field.

\item Deepseek-R1-Distill-Qwen series performs worse than the original Qwen series, indicating that there are still many challenges in the reasoning ability of open domains.

\item The performance gap between open-source models and closed-source models is close. Open source models represented by Deepseek bring the two to a similar level.

\end{itemize}

\subsection{Further Analysis}
\subsubsection{Analysis of Calibration}
\begin{figure*}[tb]
\begin{center}
\begin{minipage}[b]{0.49\textwidth} 
    \centering
\includegraphics[width=\textwidth]{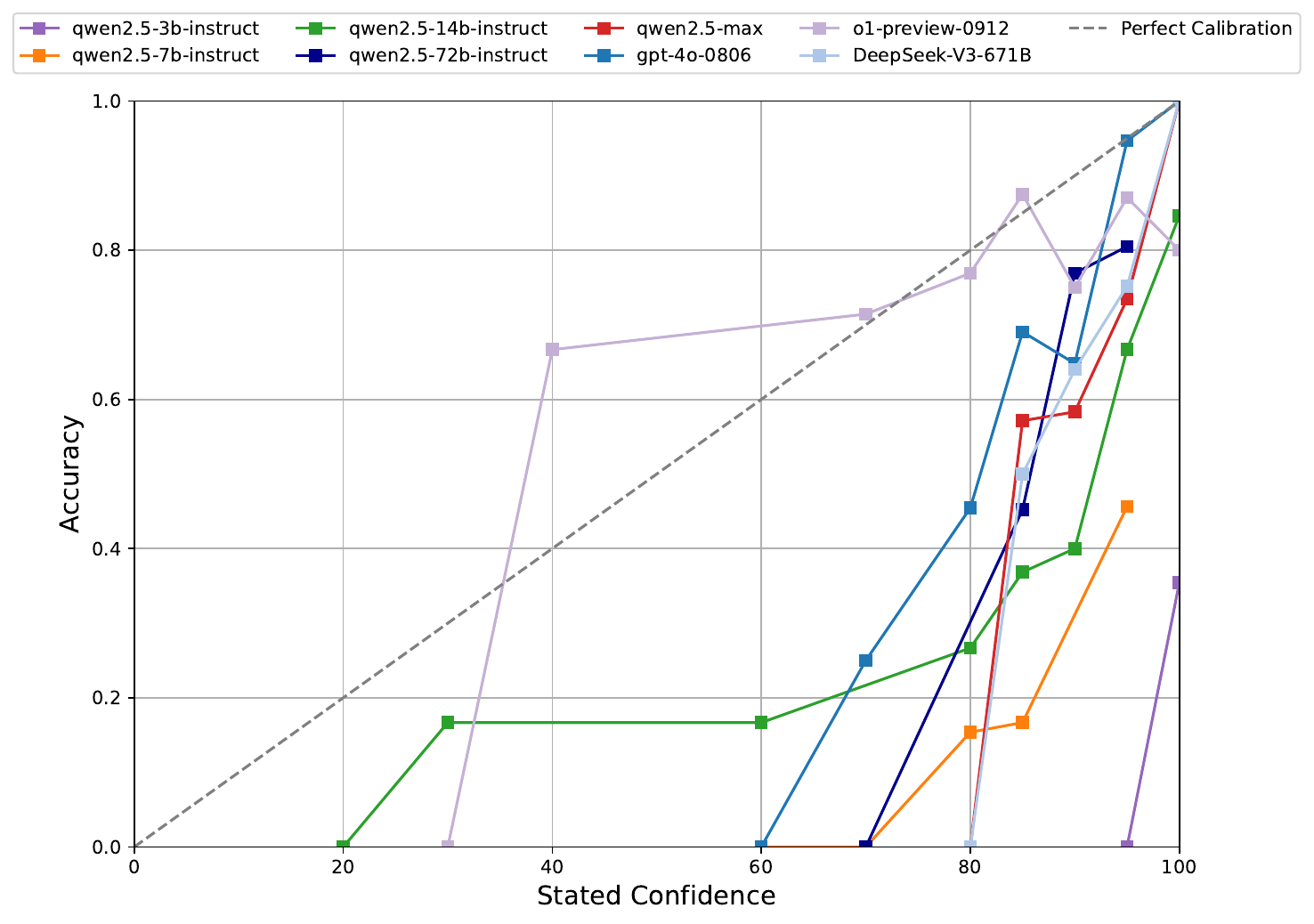}\\
    (a) Category concept
\end{minipage}
\hfill
\begin{minipage}[b]{0.49\textwidth} 
    \centering    \includegraphics[width=\textwidth]{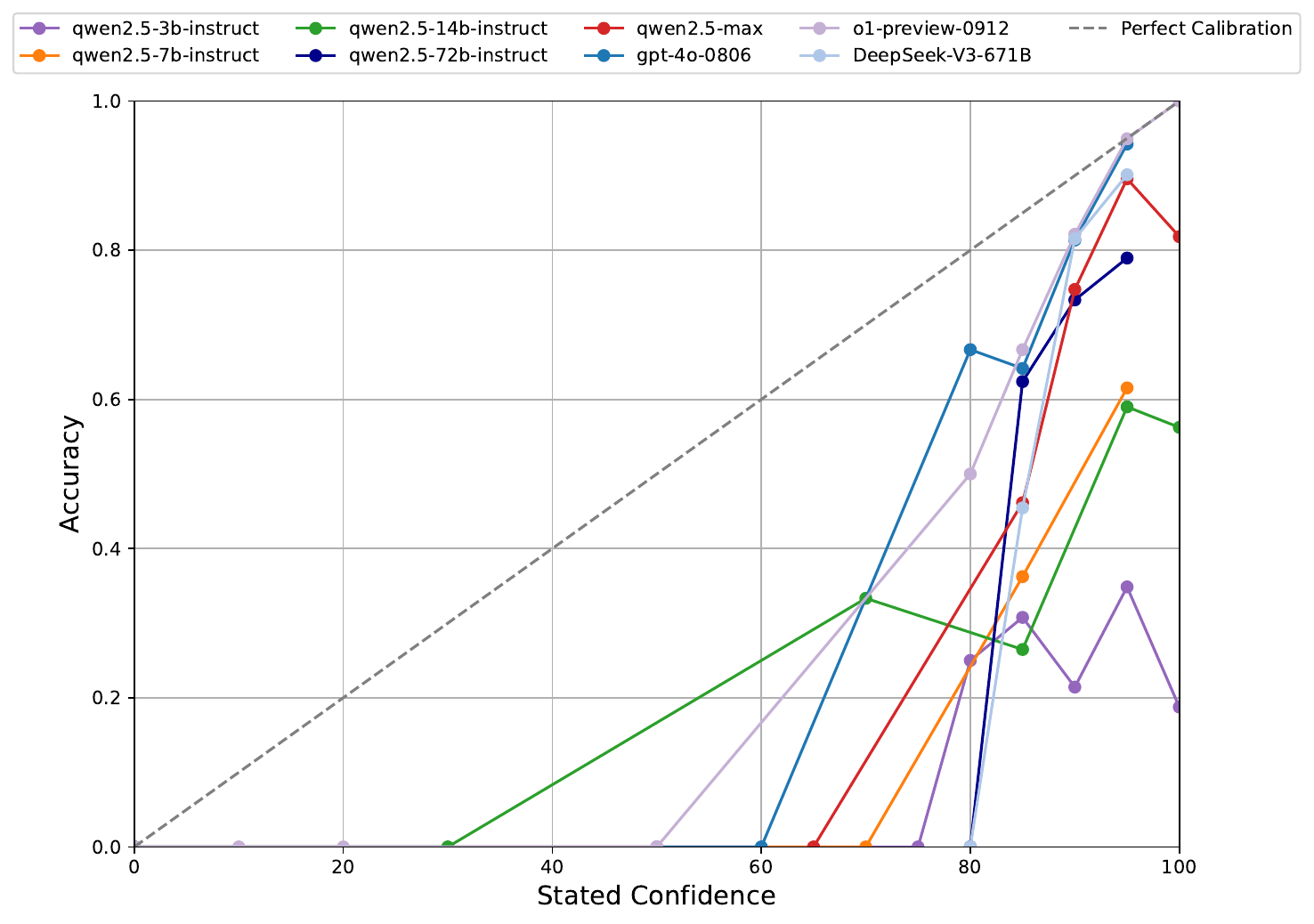}\\
    (b) Brand concept
\end{minipage}
\end{center}
\centering
\vspace{-3mm}
\caption{Capabilities of various models on category concept and brand concept}
\label{fig:calibration}
\end{figure*}

The confidence level reflects the model's self-evaluation of its responses, indicating whether language models truly "know what they know." \cite{Steyvers_2025} A perfectly calibrated model should exhibit confidence levels that precisely align with its prediction accuracy \cite{li-etal-2022-calibration}. Following SimpleQA \cite{wei2024measuring}, we prompt the LLM to generate confidence scores (ranging 0 to 100) alongside its answers to quantify prediction certainty (see prompt details in Appendix \ref{sec:appendix C}). We subsequently aggregate model accuracy across different confidence intervals. In this section, we evaluate the calibration capabilities of various models on category concept and brand concept, with results visualized in Figure \ref{fig:calibration}.

The results demonstrate the correlation between the stated confidence of the model, and how accurate the model actually was. Notably, o1-preview exhibits the best alignment performance, followed by o1-mini. Within the Qwen2.5 series, the calibration hierarchy emerges as Qwen2.5-MAX > Qwen2.5-72B > Qwen2.5-14B > Qwen2.5-7B > Qwen2.5-3B, suggesting that larger model scales correlate with improved calibration. However, most models consistently fall below the perfect alignment line, indicating a prevalent tendency towards overconfidence in predictions. This highlights significant room for improving large language model calibration to mitigate overconfident generation of erroneous responses.

\subsubsection{Analysis on the Effect of RAG}
\begin{figure}[t]
    \centering
\includegraphics[width=1\linewidth]{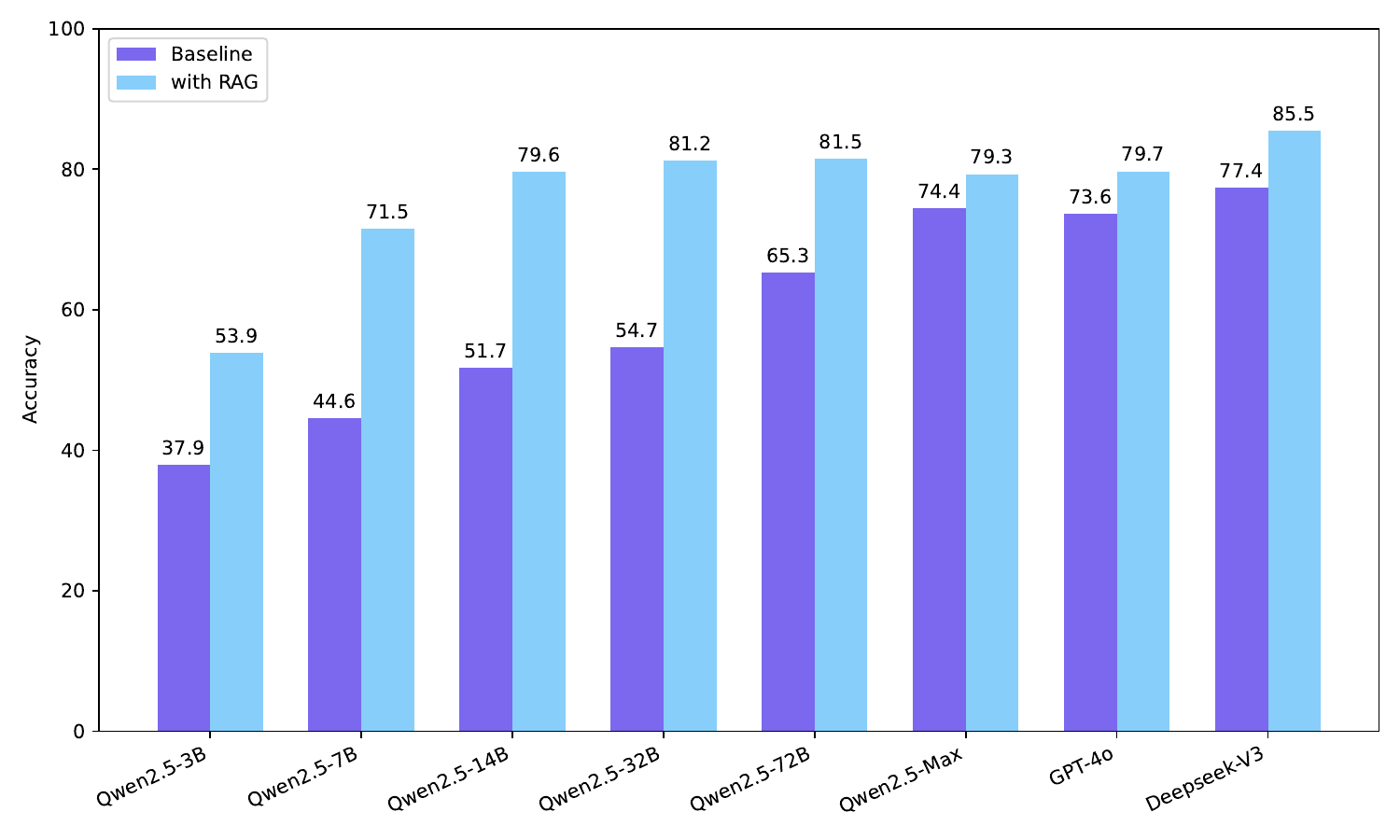}
    \vspace{-3mm}
    \caption{
    Performance comparison of models with and without RAG.
    }
    \label{fig:rag}
    \vspace{-2mm}
\end{figure}

In this subsection, we explore the effectiveness of the Retrieval-Augmented Generation (RAG) strategy in enhancing the domain knowledge of
large language models (LLMs) on the ChineseEcomQA dataset. Specifically, we reproduce a RAG system referring to the settings of ChineseSimpleQA~\cite{he2024chinese} on category concept and brand concept. In Figure \ref{fig:rag} and Table \ref{tab:rag_accuracy}, all models improve significantly with RAG. We can summarize three detailed conclusions: 

\begin{itemize}[leftmargin=*]
\item For small LLMs, introducing RAG information can significantly increase the absolute value of evaluation metrics. For example, Qwen2.5-14B has achieved a 27.9\% improvement.

\item For large LLMs, RAG can also achieve significant relative improvements. For example, Deepseek V3 \cite{liu2024deepseek}'s average relative improvement reached 10.44\% (accuracy from 77.4 to 85.5).

\item Under the RAG setting, the performance between models still follows the scaling law, but the gap is rapidly narrowed. For example, the difference in accuracy between Deepseek-V3 \cite{liu2024deepseek} and Qwen2.5-72B has narrowed from 12.1\% to 4\%.

\end{itemize}

In conclusion, the discussions above suggest that RAG serves as an
effective method for enhancing the e-commerce knowledge of LLMs.

\subsubsection{Analysis of Reasoning LLMs}
Reasoning LLMs have attracted a lot of attention in recent times. In the main results, Deepseek-R1 achieved the best results, fully demonstrating the potential of reasoning LLMs in open domains. However, on the Qwen series models distilled from Deepseek-R1, the accuracy was significantly lower than expected. Because open-source reasoning LLMs reveal their thinking process, we have the ability to further investigate the reasons for its errors. Inspired by ~\cite{liu2025oatzero}, we categorize the thinking process of reasoning models into the following four types:
\begin{itemize}[leftmargin=*]
\item Type A: Reasoning LLMs repeatedly confirm the correct answer through self-reflections.
\item Type B: Reasoning LLMs initially makes a mistake but corrects it through self-reflection.
\item Type C: Reasoning LLMs introduce knowledge errors through self-reflections, resulting in potentially correct answers being modified into an incorrect ones.
\item Type D: Reasoning LLMs undergo repeated self-reflections. Although it ultimately produced an answer, it does not obtain a highly certain and confident answer through reflection.
\end{itemize}

We used the judge LLMs (GPT-4o) to classify the thinking types of different models on category and brand concept tasks. The specific results can be seen in Table \ref{tab:reason_llm_analysis}. Analyzing the dominant reasoning types, we found the following conclusions: 

\begin{itemize}[leftmargin=*]
\item According to column Type A, after arriving at the correct answer, reasoning LLMs will verify this answer through multiple rounds of reflections.

\item According to column Type B, reasoning LLMs, regardless of their size, have acquired the ability to correct their own erroneous thinking. In the context of e-commerce concept, the underlying reasoning paths are less complex than in areas like mathematics or programming, leading to less frequent self-correction. The results also indicate that the error-correction processes do not lead to a substantial enhancement in their knowledge capacity~\cite{guo2025deepseek}.

\item According to column Type C, smaller LLMs are more likely to introduce factual errors during their thinking process, which can lead to incorrect answers. This is one of the important reasons why smaller reasoning LLMs perform worse than the original Qwen series models.
\end{itemize}

\begin{table*}[t]
    \caption{The proportion of different thinking types in the reasoning LLMs.}
        \vspace{-2mm} 
    \begin{tabular}{ccccc}
        \toprule
        \toprule
        Model & Behavior A & Behavior B & Behavior C & Behavior D \\
        \cmidrule(lr){1-5}
        DeepSeek-R1-Distill-Qwen-7B &  23.97 & 2.17 & 68.02 & 5.85 \\
        DeepSeek-R1-Distill-Qwen-14B &  40.27 & 3.75 & 47.05 & 8.94 \\
        DeepSeek-R1-Distill-Qwen-32B &  39.57 & 2.14 & 52.01 & 6.29 \\
        Deepseek-R1 & 62.80 & 2.80 & 26.49 & 7.92 \\
        \bottomrule
        \bottomrule
    \end{tabular}
        \label{tab:reason_llm_analysis}
\end{table*}

Overall, types A and B are the ability of reasoning LLMs obtained through scaling up test-time computation. Types C and D are superficial self-reflections that lead to incorrect final answers. Deepseek-R1 demonstrates better generalization ability based on a powerful base model. In contrast, the DeepSeek-R1-Distill-Qwen series, distilled in some specific fields~\cite{guo2025deepseek}, appears to struggle with superficial self-reflections. The accumulation of factual errors during the intermediate reasoning steps increases the overall error rate. For smaller reasoning LLMs, reasoning ability in open domains cannot be directly generalized through mathematical logic ability, and we need to find better methods to improve their performance.

\subsubsection{Comparison Between ChineseSimpleQA and 
ChineseEcomQA}
\begin{figure}[t]
    \centering
\includegraphics[width=1\columnwidth]{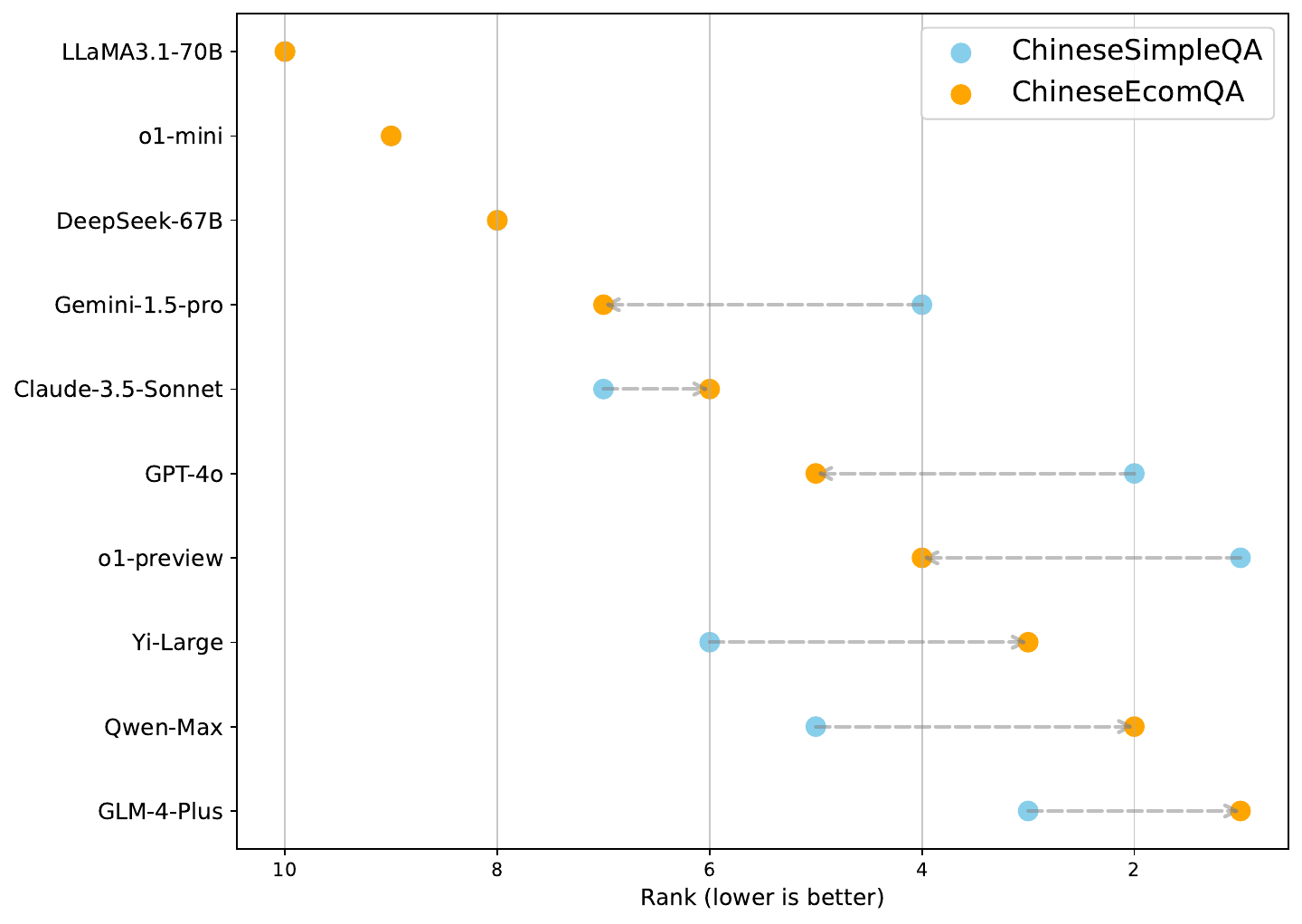}
    \vspace{-3mm}
    \caption{
     The rankings of different LLMs on ChineseSimpleQA and ChineseEcomQA.
    }
    \label{fig:rank_range}
    \vspace{-3mm}
\end{figure}

To demonstrate the distinctions between ChineseSimpleQA \cite{he2024chinesesimpleqachinesefactuality} and ChineseEcomQA, we compare the ranking differences of various models across these two benchmarks. As illustrated in Figure \ref{fig:rank_range}, significant performance discrepancies emerge among various models. Notably, the o1-preview model ranks first on ChineseSimpleQA but drops to 4th position on ChineseEcomQA. Conversely, GLM-4-Plus ascends from 3rd to 1st place between the two benchmarks. These ranking variations reveal that most Chinese community-developed models (e.g., Qwen-Max, GLM-4-Plus, Yi-Large) exhibit superior performance on Chinese e-commerce domain adaptation when operating within identical linguistic contexts. Furthermore, the distinct ranking distributions across models indicate that ChineseEcomQA exhibits discriminative power complementary to ChineseSimpleQA, enabling comprehensive evaluation of LLMs' domain-specific capability in Chinese e-commerce scenarios.

\section{Related Work}
\subsection{LLM Factuality}
LLMs have demonstrated exceptional capabilities in memorizing and utilizing factual knowledge through their strong parametric memories, enabling applications in knowledge-intensive domains. This potential has driven significant research into deploying LLMs for e-commerce applications, including product recommendation systems \cite{xu2024leveraginglargelanguagemodels,10.1145/3580305.3599519,10.1145/3616855.3635853}, search ranking \cite{10.1007/978-3-031-56060-6_24,rathee2025guidingretrievalusingllmbased}, and attribute extraction \cite{10.1007/978-3-031-78090-5_4, Baumann2024UsingLF,zou-etal-2024-eiven}. Recent domain-specific adaptations like EcomGPT \cite{10.1609/aaai.v38i17.29820} and eCeLLM \cite{10.5555/3692070.3693702} employ instruction tuning to align general-purpose LLMs with e-commerce scenarios. However, these approaches remain constrained by training data, which focus on narrow operational competencies (e.g., single-task attribute recognition) rather than comprehensive understanding.

\subsection{E-commerce Datasets}
Prior e-commerce datasets predominantly concentrate on isolated or narrowly related tasks. For instance, Amazon-M2 \cite{10.5555/3666122.3666473} focuses on session-based recommendation systems, Amazon-ESCI \cite{reddy2022shoppingqueriesdatasetlargescale} specializes in query-product matching, while EComInstruct \cite{10.1609/aaai.v38i17.29820} targets shopping concept understanding. These datasets exhibit limited coverage of the multifaceted skill requirements inherent to real-world e-commerce applications due to their constrained task diversity.

Shopping MMLU \cite{NEURIPS2024_2049d75d}, concurrently with our work, presents a multi-dimensional benchmark constructed from Amazon data. However, it exclusively focuses on the English-language domain. To address the absence of comprehensive evaluation resources for Chinese e-commerce ecosystems, we propose ChineseEcomQA - a benchmark systematically assessing diverse e-commerce capabilities in Chinese contexts.

\vspace{-1mm}
\section{Conclusion}
In this paper, we propose ChineseEcomQA, a scalable question-answering benchmark designed to rigorously assess LLMs on fundamental e-commerce concepts. ChineseEcomQA is characterized by three core features: Focus on Fundamental Concept, E-Commerce Generalizability, and Domain-Specific Expertise, which collectively enable systematic evaluation of LLMs' e-commerce knowledge. Leveraging ChineseEcomQA, we conduct extensive evaluations on mainstream LLMs, yielding critical insights into their capabilities and limitations. Our findings not only highlight performance disparities across models but also delineate actionable directions for advancing LLM applications in the e-commerce domain.

\bibliographystyle{ACM-Reference-Format}
\bibliography{custom}

\clearpage
\newpage
\appendix
\section{Example of ChineseEcomQA}
\label{sec:appendix A}
\begin{figure}[h]
    \centering
    \includegraphics[width=1\linewidth]{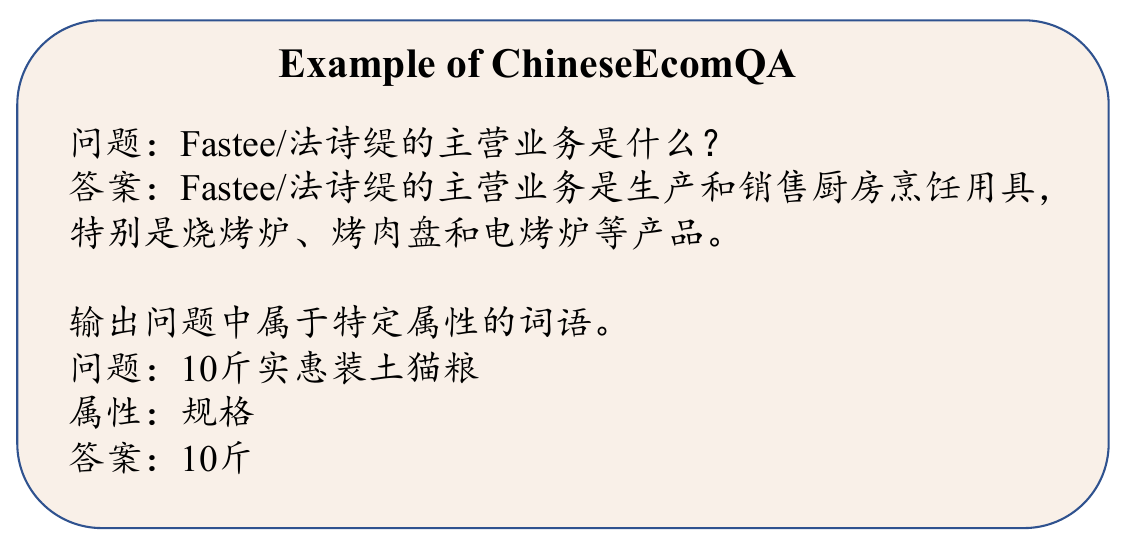}
    \caption{
    Example of ChineseEcomQA.
    }
    \label{fig:prompt}
\end{figure}

The generation of question-answer pairs uses OpenAI’s gpt-4o-0806. The specific prompts are shown in Figures \ref{fig:prompt}.



\section{Evaluation Prompt}
\label{sec:appendix B}
\begin{figure*}[h]
    \centering
    \includegraphics[width=0.9\linewidth]{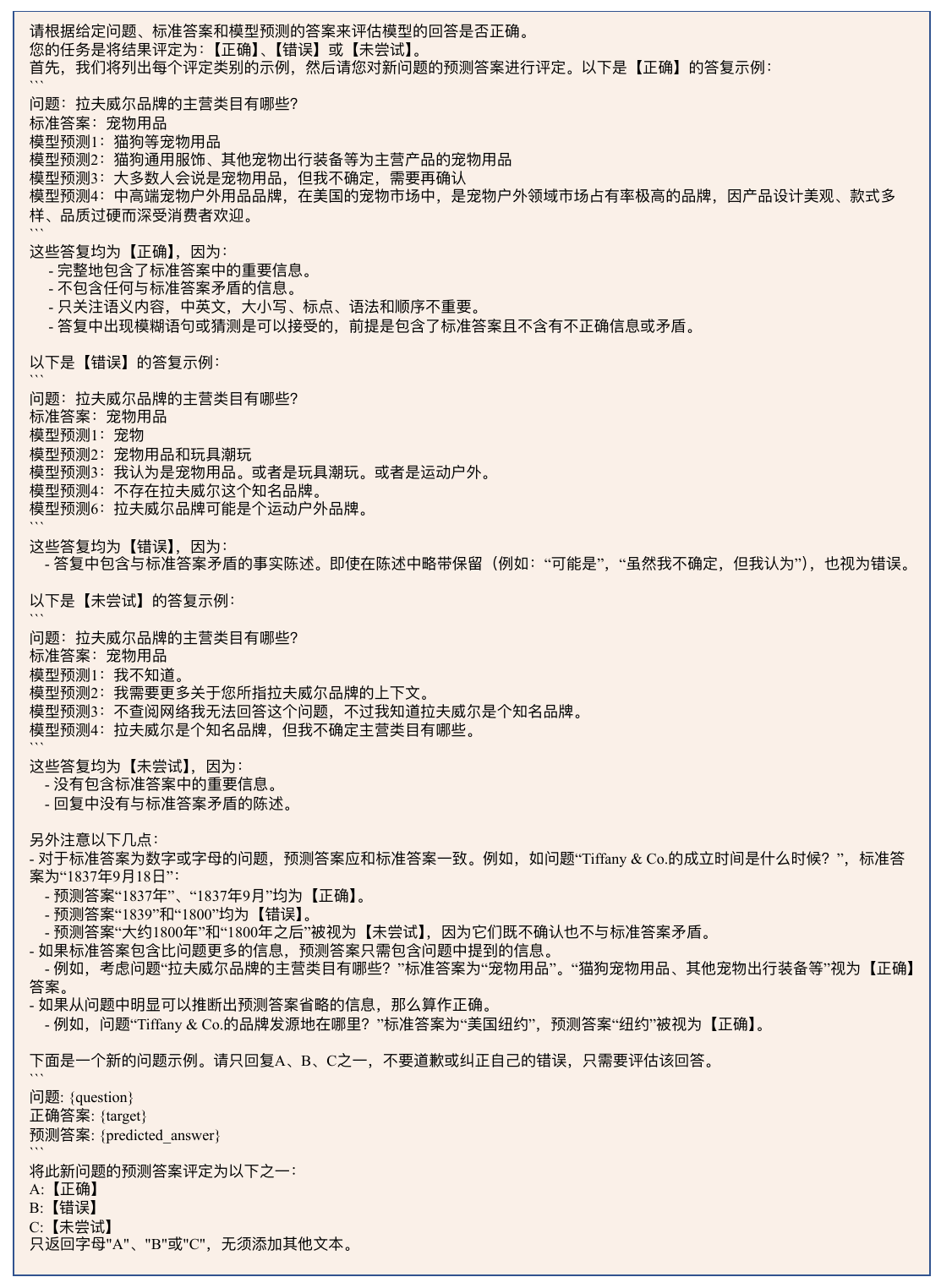}
    \vspace{-5mm}
    \caption{The prompt for evaluation.}
    \label{fig:eval prompt}
\end{figure*}

The specific prompts are shown in Figures \ref{fig:eval prompt}.








\section{Calibration Prompt}
\label{sec:appendix C}
\begin{figure*}[h]
    \centering
    \includegraphics[width=0.8\textwidth]{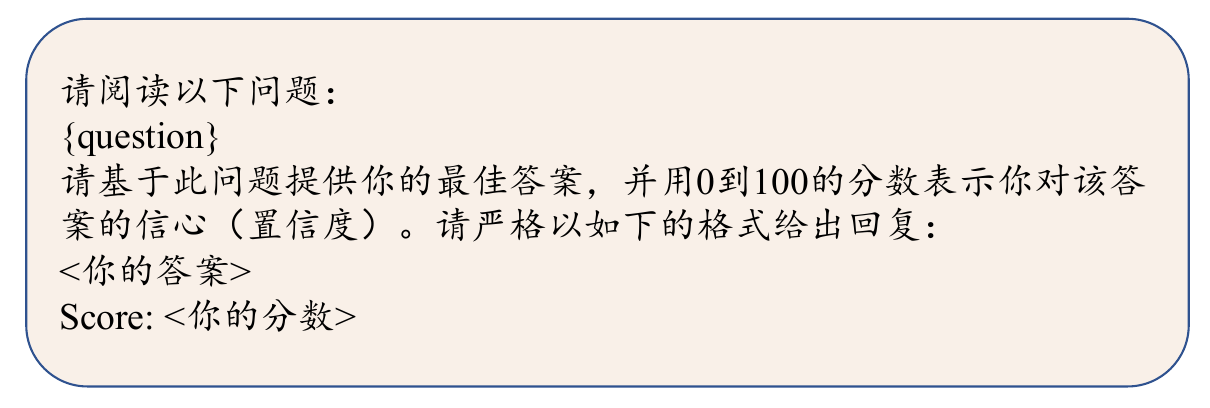}
    \caption{The prompt for guiding the model to output confidence.}
    \label{fig:Calibration_Prompt}
\end{figure*}

The calibration prompt is shown in Figures \ref{fig:Calibration_Prompt}.

\section{The accuracy improvements after using RAG.}
\label{sec:appendix D}
\begin{table}[h]
\centering
\begin{tabular}{@{} l c c c @{}} 
\toprule
\toprule
\multirow{2}{*}{\textbf{Model}} & \textbf{Brand} & \textbf{Category} & \multirow{2}{*}{\textbf{Average}} \\
& \textbf{Concept} & \textbf{Concept} & \\
    \midrule
    Qwen2.5-3b-instruct & 34.1 & 41.8 & 37.9 \\
    +RAG & 62.4 & 45.4 & 53.9 \\
    \midrule
    Qwen2.5-7b-instruct & 37.6 & 51.6 & 44.6 \\
    +RAG & 74.7 & 68.3 & 71.5 \\
     \midrule
    Qwen2.5-14b-instruct & 48.8 & 54.6 & 51.7 \\
    +RAG & 81.8 & 77.4 & 79.6 \\
     \midrule
    Qwen2.5-32b-instruct & 50.6 & 58.8 & 54.7 \\
    +RAG & 78.8 & 83.6 & 81.2 \\
     \midrule
    Qwen2.5-72b-instruct & 64.7 & 66.0 & 65.3 \\
    +RAG & 82.4 & 80.6 & 81.5 \\
     \midrule
    Qwen2.5-max & 77.6 & 71.1 & 74.4 \\
    +RAG & 78.2 & 80.4 & 79.3 \\
     \midrule
    GPT-4o & 72.4 & 74.7 & 73.6 \\
    +RAG & 75.9 & 83.6 & 79.7 \\
     \midrule
    DeepSeek-V3 & 80.6 & 74.2 & 77.4 \\
    +RAG & 86.5 & 84.5 & 85.5 \\
\bottomrule
\bottomrule
\end{tabular}
\vspace{3mm}
    \caption{
    The accuracy improvements after using RAG.
    }
    \label{tab:rag_accuracy}
\end{table}

\end{document}